\documentclass{article}
\usepackage[utf8]{inputenc}
\usepackage{main}
\usepackage{microtype}
\usepackage{graphicx}
\usepackage{subfig}
\usepackage{times}
\usepackage{latexsym}
\usepackage{amsmath}
\usepackage{float}
\usepackage{footnote}
\usepackage{enumitem}
\usepackage{bm}
\usepackage{arydshln}
\usepackage{booktabs}
\usepackage{multicol}
\usepackage{multirow}
\usepackage{color}
\usepackage{xcolor}     
\usepackage{colortbl}
\usepackage{bbding}
\usepackage{makecell}
\usepackage{mathtools}
\usepackage{imakeidx}
\makeindex
\usepackage{arydshln}
\usepackage{lipsum}
\usepackage{natbib}
\usepackage[toc]{multitoc}
\usepackage[edges]{forest}
\usepackage[normalem]{ulem}
\definecolor{mydarkblue}{rgb}{0,0.08,0.45}
\usepackage[colorlinks=true,linkcolor=mydarkblue,citecolor=mydarkblue,filecolor=mydarkblue,urlcolor=mydarkblue]{hyperref}
\usepackage{caption}
\usepackage{CJKutf8}
\usepackage{awesomebox} 
\usepackage{bbding}
\usepackage[most]{tcolorbox}

\usepackage[tikz]{bclogo}
\usepackage[framemethod=tikz]{mdframed}
\definecolor{bgblue}{RGB}{245,243,253}
\definecolor{ttblue}{RGB}{91,194,224}

\mdfdefinestyle{mystyle}{%
  rightline=true,
  innerleftmargin=10,
  innerrightmargin=10,
  outerlinewidth=3pt,
  topline=false,
  rightline=true,
  bottomline=false,
  skipabove=\topsep,
  skipbelow=\topsep
}

\newtcolorbox{myboxi}[1][]{
  breakable,
  title=#1,
  colback=red!5,
  colbacktitle=red!5,
  coltitle=black,
  fonttitle=\bfseries,
  bottomrule=0pt,
  toprule=0pt,
  leftrule=2pt,
  rightrule=2pt,
  titlerule=0pt,
  arc=0pt,
  outer arc=0pt,
  colframe=red,
}

\newtcolorbox{myboxnote}[1][]{
  breakable,
  title=#1,
  colback=orange!0,
  colbacktitle=orange!0,
  coltitle=black,
  fonttitle=\bfseries,
  bottomrule=0pt,
  toprule=0pt,
  leftrule=2pt,
  rightrule=2pt,
  titlerule=0pt,
  arc=0pt,
  outer arc=0pt,
  colframe=orange,
}

\newtcolorbox{myboxii}[1][]{
  breakable,
  freelance,
  title=#1,
  colback=white,
  colbacktitle=white,
  coltitle=black,
  fonttitle=\bfseries,
  bottomrule=0pt,
  boxrule=0pt,
  colframe=white,
  overlay unbroken and first={
  \draw[red!75!black,line width=3pt]
    ([xshift=5pt]frame.north west) -- 
    (frame.north west) -- 
    (frame.south west);
  \draw[red!75!black,line width=3pt]
    ([xshift=-5pt]frame.north east) -- 
    (frame.north east) -- 
    (frame.south east);
  },
  overlay unbroken app={
  \draw[red!75!black,line width=3pt,line cap=rect]
    (frame.south west) -- 
    ([xshift=5pt]frame.south west);
  \draw[red!75!black,line width=3pt,line cap=rect]
    (frame.south east) -- 
    ([xshift=-5pt]frame.south east);
  },
  overlay middle and last={
  \draw[red!75!black,line width=3pt]
    (frame.north west) -- 
    (frame.south west);
  \draw[red!75!black,line width=3pt]
    (frame.north east) -- 
    (frame.south east);
  },
  overlay last app={
  \draw[red!75!black,line width=3pt,line cap=rect]
    (frame.south west) --
    ([xshift=5pt]frame.south west);
  \draw[red!75!black,line width=3pt,line cap=rect]
    (frame.south east) --
    ([xshift=-5pt]frame.south east);
  },
}

\usepackage{fancyhdr} 
\usepackage{blindtext} 

\DeclareCaptionFont{black}{\color{black}}

\definecolor{myblue}{rgb}{0.9, 0.1, 0.94}
\definecolor{mygreen}{rgb}{0.64, 0.56, 0.88}
\definecolor{myyellow}{rgb}{0.68, 0.6, 0.1}
\definecolor{fancygreen}{rgb}{0.33, 0.68, 0.20}
\definecolor{salmon}{rgb}{0.94, 0.52, 0.49}
\definecolor{tablegreen}{rgb}{0.82, 0.94, 0.75}
\definecolor{tableblue}{rgb}{0.81, 0.90, 0.94}
\definecolor{tablered}{rgb}{0.97, 0.85, 0.85}
\definecolor{tableorange}{rgb}{0.96, 0.85, 0.81}

\definecolor{myorange}{rgb}{1.0, 0.49, 0.0}	
\definecolor{tlgreen}{rgb}{0.33, 0.68, 0.20}

\newenvironment{itemize*}%
 {\leftmargini=10pt\begin{itemize}%
  \setlength{\itemsep}{0pt}%
  \setlength{\parskip}{0pt}%
  }%
 {\end{itemize}}
\newenvironment{enumerate*}%
 {\begin{enumerate}%
  \setlength{\itemsep}{0pt}%
  \setlength{\parskip}{0pt}}%
 {\end{enumerate}}

\tikzset{%
    parent/.style =          {align=center,text width=2cm,rounded corners=3pt, line width=0.3mm, fill=gray!10,draw=gray!80},
    child/.style =           {align=center,text width=2.3cm,rounded corners=3pt, fill=blue!10,draw=blue!80,line width=0.3mm},
    grandchild/.style =      {align=center,text width=2cm,rounded corners=3pt},
    greatgrandchild/.style = {align=center,text width=1.5cm,rounded corners=3pt},
    greatgrandchild2/.style = {align=center,text width=1.5cm,rounded corners=3pt},    
    referenceblock/.style =  {align=center,text width=1.5cm,rounded corners=2pt},
    pretrain/.style =           {align=center,text width=1.8cm,rounded corners=3pt, fill=blue!10,draw=blue!80,line width=0.3mm},   
    pretrain_work/.style =           {align=center, text width=5cm,rounded corners=3pt, fill=blue!10,draw=blue!0,line width=0.3mm},  
    template/.style =           {align=center,text width=1.8cm,rounded corners=3pt, fill=red!10,draw=red!80,line width=0.3mm},   
    template_work/.style =           {align=center,text width=5cm,rounded corners=3pt, fill=red!10,draw=red!0,line width=0.3mm},    
    answer/.style =           {align=center,text width=1.8cm,rounded corners=3pt, fill= cyan!10,draw= cyan!80,line width=0.3mm},   
    answer_work/.style =           {align=center,text width=5cm,rounded corners=3pt, fill= cyan!10,draw= cyan!0,line width=0.3mm},      
    multiple/.style =           {align=center,text width=1.8cm,rounded corners=3pt, fill= orange!10,draw= orange!80,line width=0.3mm},   
    multiple_work/.style =           {align=center,text width=5cm,rounded corners=3pt, fill= orange!10,draw= orange!0,line width=0.3mm},        
    tuning/.style =           {align=center,text width=1.8cm,rounded corners=3pt, fill= magenta!10,draw= magenta!80,line width=0.3mm},   
    tuning_work/.style =           {align=center,text width=5cm,rounded corners=3pt, fill= magenta!10,draw= magenta!0,line width=0.3mm},          
}

\usepackage{minitoc}
\usepackage[toc,page,header]{appendix}
\usepackage{soul}
\usepackage{makecell}
\usepackage{framed,multirow}
\usepackage{threeparttable}
\usepackage{amssymb}
\usepackage{pifont}
\usepackage{algorithm}
\usepackage{layouts}
\usepackage{listings}
\usepackage{colortbl}
\usepackage{pythonhighlight}
\usepackage{tikz}
\usepackage{enumitem}
\usepackage{mathrsfs}

\definecolor{iblue}{rgb}{0.06, 0.75, 1.0}
\definecolor{igray}{rgb}{0.00, 0.00, 0.00}

\newcolumntype{P}[1]{>{\centering\arraybackslash}p{#1}}
\newlength\savewidth
\newcommand{\cmark}{\ding{51}}%
\newcommand{\xmark}{\ding{55}}%

\definecolor{mygray}{gray}{0.6}
\definecolor{mygray-bg}{gray}{0.9}

\definecolor{qixin_red}{RGB}{227,36,43}
\definecolor{qixin_green}{RGB}{60,186,84}

\usepackage{etoolbox}
\usepackage{natbib}
\usepackage{url}
\newcounter{bibcount}
\makeatletter
\patchcmd{\@lbibitem}{\item[}{\item[\hfil\stepcounter{bibcount}{[\thebibcount]}}{}{}
\renewcommand\NAT@bibsetup%
  [1]{\setlength{\leftmargin}{\bibhang}\setlength{\itemindent}{-\parindent}%
      \setlength{\itemsep}{\bibsep}\setlength{\parsep}{\z@}}
\makeatother

\begin{document}

\title{Synthetic Data as Validation}

\author{
    Qixin Hu \\
    The Chinese University of Hong Kong \\
\texttt{qixinhu@cuhk.edu.hk}
\And
    Alan Yuille \\
    Johns Hopkins University \\
    \texttt{ayuille1@jhu.edu}
\And
    Zongwei Zhou\thanks{Correponding author} \\
    Johns Hopkins University \\
    \texttt{zzhou82@jh.edu} \\
}
  
\maketitle

\begin{abstract}

This study leverages synthetic data as a validation set to reduce overfitting and ease the selection of the best model in AI development. While synthetic data have been used for augmenting the training set, we find that synthetic data can also significantly diversify the validation set, offering marked advantages in domains like healthcare, where data are typically limited, sensitive, and from out-domain sources (i.e., hospitals). In this study, we illustrate the effectiveness of synthetic data for early cancer detection in computed tomography (CT) volumes, where synthetic tumors are generated and superimposed onto healthy organs, thereby creating an extensive dataset for rigorous validation. Using synthetic data as validation can improve AI robustness in both in-domain and out-domain test sets. Furthermore, we establish a new continual learning framework that continuously trains AI models on a stream of out-domain data with synthetic tumors. The AI model trained and validated in dynamically expanding synthetic data can consistently outperform models trained and validated exclusively on real-world data. Specifically, the DSC score for liver tumor segmentation improves from 26.7\% (95\% CI: 22.6\%--30.9\%) to 34.5\% (30.8\%--38.2\%) when evaluated on an in-domain dataset and from 31.1\% (26.0\%--36.2\%) to 35.4\% (32.1\%--38.7\%) on an out-domain dataset. Importantly, the performance gain is particularly significant in identifying very tiny liver tumors (radius $<$ 5mm) in CT volumes, with Sensitivity improving from 33.1\% to 55.4\% on an in-domain dataset and 33.9\% to 52.3\% on an out-domain dataset, justifying the efficacy in early detection of cancer. The application of synthetic data, from both training and validation perspectives, underlines a promising avenue to enhance AI robustness when dealing with data from varying domains. As open science, we have released the codes and models at \href{https://github.com/MrGiovanni/SyntheticValidation}{https://github.com/MrGiovanni/SyntheticValidation}.

\end{abstract}


\clearpage

\section{Introduction}\label{sec:introduction}

Standard AI development divides the dataset into a training set and a test set; the former is used for model training and the latter for evaluation~\citep{russell2010artificial,gareth2013introduction}. The AI model is updated every \textit{training epochs}, resulting in a number of intermediate models during the training trajectory. The performance of these models tends to improve on the training set, but this does not mean that the performance on the test set also improves due to the over-fitting problem~\citep{kuhn2013applied}. A question then arises: \textit{How do we identify the best model that performs well on the test set, especially when it is evaluated on test sets taken from different domains?} A prevalent strategy is to delineate a validation set from the training set~\citep{ripley2007pattern}. This validation set neither contributes to training nor to evaluating the AI performance. Instead, it functions as an independent set to fix the training hyper-parameters and, more importantly, to estimate the performance of each model on different datasets, thus enabling the selection of the best model from the many intermediate models during the training trajectory. 

The validation set is often kept small. Naturally, we would like to maximize the use of the training data. Annotating data for AI training is time-consuming and expensive, requiring specialized expertise, so the annotated datasets are limited in size in many fields~\citep{zhou2021towards}. Allocating too many annotated data for validation would inevitably diminish the training set size and compromise the AI training. On the other hand, the validation set should be sufficiently representative to provide a reliable performance estimate on unseen data. An overly small validation set might risk the reliability of performance estimation and checkpoint selection. As a result, the calibration of the validation set remains largely empirical and lacks systematic investigation for better alternatives to select the best checkpoint. Fulfilling this knowledge gap is particularly important in scenarios where real-world data are scarce, sensitive, or costly to collect and annotate, as seen in the field of \textit{AI for healthcare}~\citep{zhou2022interpreting}. Therefore, our study uses early detection of cancerous tumors in computed tomography (CT) volumes as a demonstration. While early detection of cancer holds immense clinical potential, it faces profound constraints like disease prevalence and annotation difficulty to collect examples of early-stage tumors~\citep{crosby2022early}. The scarcity of annotated early cancer not only constrains the data available for validation but also amplifies the overfitting problem inherent in a small, biased validation set, potentially causing underdiagnosis and overdiagnosis.

We propose using synthetic data as validation, a strategy that guarantees the full utilization of the training set while ensuring ample data diversity for validation. Data synthesis has held longstanding interest and presents numerous intriguing merits for augmenting training and test data~\citep{hu2023label,gao2023synthetic} as reviewed in \S\ref{sec:related_work}, but its use in validation has seldom been explored. We find that synthetic data can facilitate a more reliable performance estimate on unseen data and effectively address the constraints commonly associated with small, biased validation sets. Specifically, we synthesize tumors in the healthy liver, which gives us orders of magnitude larger datasets for training. To ensure the realism of the synthetic tumors, we employ a modeling-based strategy~\citep{hu2023label} to simulate cancerous tumors with controlled shape, size, texture, location, and intensity. The use of diverse, healthy CT volumes, supplemented with synthetic tumors, as validation has demonstrated efficacy in mitigating model overfitting and enhancing the selection of checkpoints. Furthermore, we relieve the pressing demand for human annotations to train AI models by utilizing CT volumes with synthetic tumors as the training set. We then assess the model's performance using a substantial number of publicly available, fully-annotated CT volumes with real-world cancerous tumors, showing that our models generalize well to these volumes from different hospitals and accurately segment the tumors at their early stage. 
Our findings can be summarized as follows:

\begin{enumerate}[leftmargin=*]
    
    \item The best model checkpoint, selected by standard AI development with an in-domain real-tumor validation set, may not necessarily be generalized to unseen data, especially for an out-domain test set. This limitation arises from the validation set failing to adequately represent corner cases.
    
    \item The best model checkpoint, selected by our strategy with a diverse synthetic-tumor validation set, tends to be generalized well to unseen data. This is because the validation set can cover theoretically infinite examples of possible cancerous tumors across diverse conditions.

    \item We introduce a novel continual learning framework. This framework integrates a continuous stream of synthetic data, characterized by diverse data distribution, for both training and validation. Traditional validation sets, constrained by static and limited in-domain real tumors, fall short in such a setting, whereas our synthetic tumors can be dynamically tailored to align with emerging distributions. Importantly, our framework can continuously generate tumors spanning a spectrum of sizes---from small to large---enhancing the detection rate of tumors at their early stages.

\end{enumerate}

Although our study focuses on AI in healthcare, the insight should be pertinent to various imaging applications within the field of computer vision. However, at the time this paper is written, very few studies in computer vision have provided evidence that training \textit{exclusively} on generated synthetic data can match or surpass the performance achieved when trained on real data~\citep{black2023bedlam}. In specific applications, integrating synthetic data with real data---essentially acting as data augmentation---has been found empirically to boost AI performance~\citep{mu2020learning,luzi2022boomerang,azizi2023synthetic,burg2023data}. In this regard, data synthesis---cancerous tumor synthesis in particular---in medical imaging is relatively more successful\footnote{The greater success of data synthesis in medical imaging (reviewed in \S\ref{sec:related_work}), compared with computer vision, can be attributed to two factors from our perspective. Firstly, the focus is primarily on synthesizing tumors rather than other components of the human anatomy. Secondly, the synthesis of tumors in 3D medical images is less complex as it does not require considerations for intricate variables such as lighting conditions, pose, and occlusion, which are typical in computer vision tasks.} with specific applications benefiting more from training exclusively on synthetic data than real data.

\section{Related Work}\label{sec:related_work}

\textbf{The dilemma of validation.}

In the field of machine learning, it is customary to use finite, static datasets with a pre-defined data split. While this standard offers a fair benchmark for comparing different AI models, it does not accurately represent real-world learning conditions. Two more realistic scenarios often arise in practice. 

\begin{itemize}[leftmargin=*]
    
    \item The first scenario is the \textit{small data regime}, commonly observed in medical applications due to constraints like disease prevalence and annotation difficulty~\citep{liu2022oard}. In such cases, curating an appropriate validation set poses a conundrum. A large validation set would compromise the size of the training set, whereas a small one may not sufficiently estimate the model's performance. Despite its critical importance, this issue has yet to receive adequate attention in the field.
    
    \item The second scenario involves dealing with \textit{a stream of data}, in a context of continual learning where the model encounters a continuous flow of new data~\citep{purushwalkam2022challenges}. A finite, static validation set proves unsuitable as it cannot accurately assess the model's capability in processing an extensive and diverse data range. We argue that a validation set---made up of real-world data---might not be needed during the training stage in such situations. Given the vastness of the training data, overfitting can be naturally avoided. Consequently, selecting the last-epoch model checkpoint could be a judicious choice.

\end{itemize}

\textbf{Progresses in data synthesis.} Real-world data often encounters challenges such as poor quality, limited quantity, and inaccessibility~\citep{zhu2022assembling,qu2023annotating,kang2023label,liu2023clip}. To tackle these obstacles, the notion of \textit{synthetic data} has emerged as a practical alternative, allowing for the generation of samples as needed~\citep{jordon2018pate,yoon2019time,chen2021synthetic}. This approach has proven valuable in addressing data limitations and facilitating machine learning processes, including computer vision~\citep{chen2019learning,ramesh2021zero}, natural language processing~\citep{collobert2008unified,brown2020language}, voice~\citep{oord2016wavenet}, and many other fields~\citep{wiese2020quant,jin2018junction, zheng2023toward}. In the medical domain, the practice of data synthesis---\textit{tumor synthesis} in particular---endeavors to produce artificial tumors in the image, which can significantly diversify the data and annotations for AI training~\citep{xing2023less} and, arguably, can strengthen the AI robustness evaluation using corner cases generated by data synthesis. Successful works related to tumor synthesis include polyp detection from colonoscopy videos~\citep{shin2018abnormal}, COVID-19 detection from Chest CT and X-ray~\citep{yao2021label,lyu2022pseudo, gao2023synthetic}, diabetic lesion detection from retinal images~\citep{wang2022anomaly}, cancer detection from fluorescence microscopy images~\citep{horvath2022metgan}, brain tumor detection from MRI~\citep{wyatt2022anoddpm}, early pancreatic cancer localization from CT~\citep{li2023early}, and pneumonia detection from ultrasound~\citep{yu2023good}. A recent study~\citep{hu2022synthetic,hu2023label} indicated that AI trained \textit{exclusively} on synthetic tumors can segment liver tumors with accuracy comparable to that on real tumors.

To the best of our knowledge, data synthesis has been widely recognized for its contribution to enhancing training and test datasets~\citep{johnson2017clevr,jordon2022synthetic,liu2023poseexaminer}, but its capacity for improving the validation set remains largely untapped. In this paper, we extend the application of synthetic data to the validation set, enabling the full use of the annotated data for AI training while ensuring diverse and comprehensive validation data in the framework of continual learning.

\begin{figure}[h]
    \centering
    \includegraphics[width=0.9\textwidth]{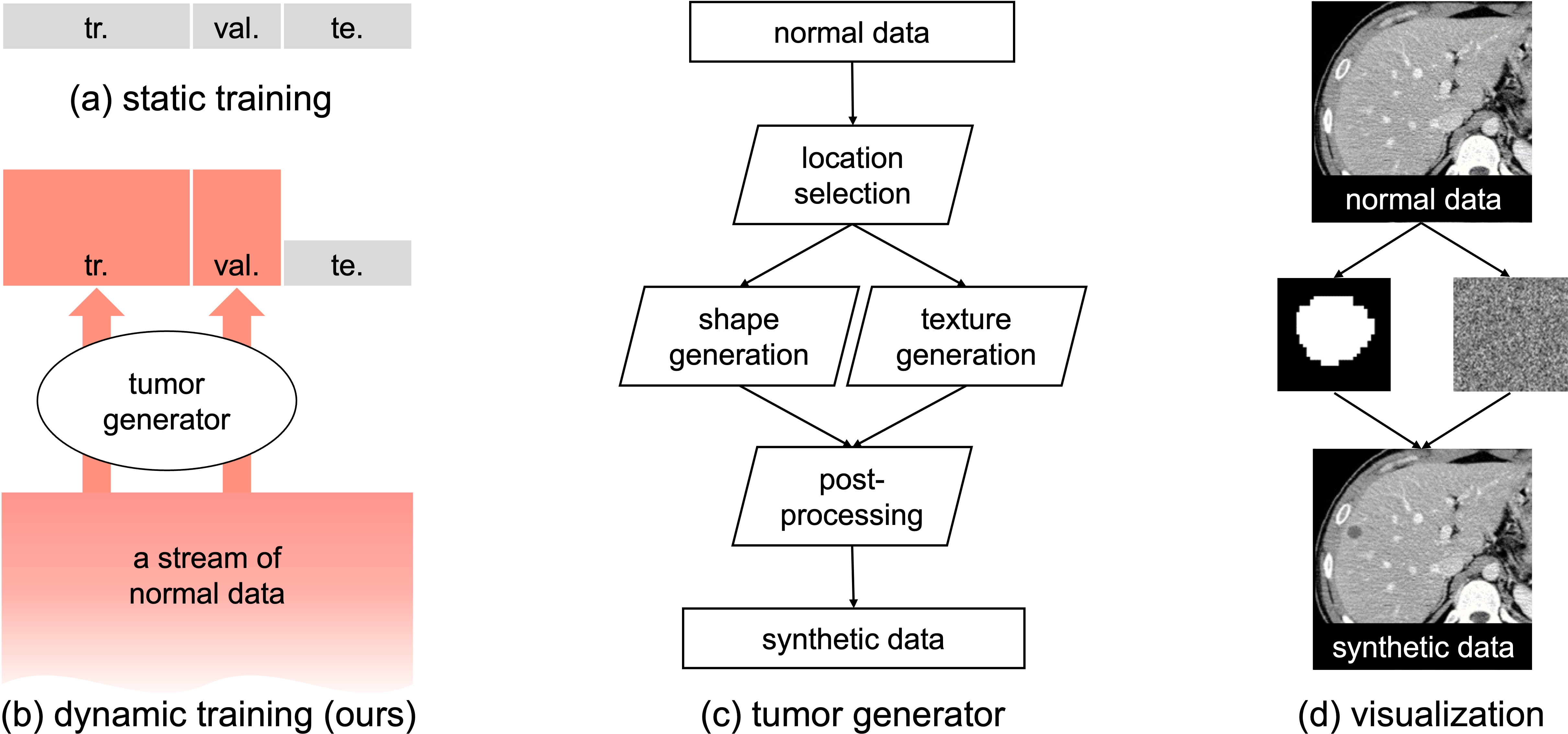}
    \caption{\textbf{Illustration of continual learning framework.} \textbf{(a)} The setting for the static training, where the real-tumor dataset is partitioned into training and validation sets. The AI model is then developed using these datasets and subsequently tested with unseen data. \textbf{(b)} Dynamic training setting integrated with synthetic data. \textbf{(c)} Tumor generator pipeline. By leveraging an advanced tumor generator, we can create a dynamic training and validation set. Models developed using the continual learning framework exhibit superior performance compared to static training settings. \textbf{(d)} Visualization of synthetic data.}
    \label{fig:methods}
\end{figure}

\section{Method \& Material}\label{sec:method}

\subsection{Continual Learning for Tumor Segmentation}\label{sec:lifelong_learning}

According to~\citet{van2019three, van2022three}, continual learning can be categorized into three settings: class-incremental learning, task-incremental learning, and domain-incremental learning. In the domain-incremental setting, which is relevant to our situation, the task remains the same while the data distribution changes. More specifically, the model sequentially encounters data from a continuum of domains (datasets):
\begin{equation}
    \label{eq:domain-incremental}
    \{X_1, Y_1\},\{X_2, Y_2\}, ... , \{X_N, Y_N\},
\end{equation}
The objective is to train a model $\mathscr{F}: X \rightarrow Y$ that can be effectively queried at any given time, regardless of the data distribution. In liver tumor segmentation tasks, $X$ is the CT volume and $Y$ is the tumor mask. The continuum of domains refers to CT volumes taken from different medical centers. 

In the setting of \textit{static} training, shown in \figureautorefname~\ref{fig:methods}(a), the AI model is trained and validated on fixed subsets of a dataset. This setting presents three limitations: Firstly, the limited scales and acquisition sources of the data, coupled with unchanged data distribution, pose challenges in generalizing to out-domain data. Secondly, the task of data annotation, specifically tumor annotation, is exceptionally challenging as it often requires the use of corroborative pathology reports. This requirement adds to the difficulty of extending the dataset. Thirdly, there are specific cases, such as extremely small tumors, where obtaining real data becomes significantly challenging. As a result, static training will likely result in biased, sub-optimal performance on unseen data, especially for out-domain test sets.

In contrast, the setting of \textit{dynamic} training achieved by synthetic data, shown in \figureautorefname~\ref{fig:methods}(b), can overcome the aforementioned limitations. In this setting, the AI model is trained and validated on a dynamically changing dataset. In our study, this dynamic dataset is a stream of normal CT volumes---over 40 million CT volumes in the United States each year. Generating synthetic tumors is advantageous because, firstly, acquiring healthy CT volumes is much easier than obtaining those with cancerous tumors. As a result, our continual learning framework can start from a diverse dataset comprising CT volumes of healthy subjects from multiple domains. Secondly, by controlling the parameters within our framework, we have the ability to generate synthetic data that fulfills specific requirements, including those of a tiny radius ($<$5 mm, shown in \appendixautorefname~\figureautorefname~\ref{fig:appendix_shape_examples}). Consequently, our framework achieves a noteworthy level of diversity, encompassing a wide array of variations. Therefore, the AI model developed using this framework of synthetic data has the potential to improve its performance on out-domain data.

\subsection{Modeling-Based Synthetic Tumor Generation}\label{sec:tumor_generation}

Following the standardized clinical guidance and statistical distribution of real tumors, as detailed in \appendixautorefname~\figureautorefname~\ref{fig:appendix_distribution}~\ref{fig:appendix_size_dis}, we develop a modeling-based strategy to generate synthetic tumors. For example, according to the Liver Imaging Reporting and Data System (LI-RADS)~\citep{chernyak2018liver}, the malignancy of hepatocellular carcinomas is determined by shape, size, location, and texture, enhancing capsule appearance. We use a sequence of morphological image-processing operations to model real tumors, as shown in \figureautorefname~\ref{fig:methods}(c). The tumor generator consists of four steps: (1) location selection, (2) shape generation, (3) texture generation, and (4) post-processing.

\begin{enumerate}[leftmargin=*]

    \item \textbf{Location selection.} Liver tumors generally do not allow the passage of preexisting blood vessels from the host tissue through them. To address this concern, we initially perform voxel value thresholding for vessel segmentation~\citep{gonzalez2009digital}. Utilizing the vessel mask acquired from this step enables us to identify if a particular location can cause the tumor-blood collision.
    
    \item \textbf{Shape generation.} Based on clinical knowledge, a tumor is initiated from a malignant cell and gradually proliferates and expands, resulting in a nearly spherical shape for small tumors ($\le$5mm). On the other hand, statistical distributions of real liver tumors indicate that larger tumors tend to exhibit an elliptical shape. This observation has inspired us to generate a tumor-like shape using an ellipsoid $\textit{ellip}(a, b, c)$, where $a, b, c$ are the lengths of the semi-axes. Additionally, we utilize elastic deformation~\citep{ronneberger2015u} to enhance the authenticity of the generated tumor shapes $D (\textit{ellip}(a, b, c), \sigma_d)$, where $\sigma_d$ control the magnitude of displacements. We show examples of the generated tumor shapes in \appendixautorefname~\figureautorefname~\ref{fig:appendix_shape_examples}.
    
    \item \textbf{Texture generation.} The generation of textures is a significant challenge due to the varied patterns found in tumors. Our current understanding of tumor textures is derived solely from clinical expertise, which considers factors such as the attenuation value and the distribution characteristics. To achieve the desired texture, we introduce Gaussian noise $ \mathcal{N}(\mu, \sigma_g) $ with a predetermined mean attenuation value, matching the standard deviation of liver tumors. Subsequently, we use cubic interpolation to smooth the texture. Furthermore, to better replicate textures obtained from CT imaging, we use a final step of texture blurring. Examples of the generated texture can be found in \appendixautorefname~\figureautorefname~\ref{fig:appendix_texture_examples}.
    
    \item \textbf{Post-processing.} The post-processing involves evaluating image characteristics through visual inspection and feedback from medical professionals. The purpose of these steps is to replicate the phenomena of mass effect and the appearance of a capsule~\citep{lee2004triple}. Mass effect refers to the phenomenon wherein the tumor undergoes growth, resulting in the displacement and deformation of surrounding tissues. We utilize local scaling warping~\citep{glasbey1998review} to replicate this effect. Additionally, we brighten the edges of the tumor, thereby simulating the capsule appearance. Consequently, CT volumes with synthetic tumors can be used for the continual learning framework, where examples of the generated liver tumors can be found in \appendixautorefname~\figureautorefname~\ref{fig:appdix_examples}.

\end{enumerate}

\section{Experiment}\label{sec:experiment}

\subsection{Dataset \& Benchmark}\label{sec:dataset_benchmark}

\begin{table*}[h]
    \footnotesize
    \centering
    \caption{
        \textbf{Datasets description.} The LiTS dataset was used to train, validate, and evaluate AI models in segmenting liver tumors. The FLARE'23 dataset was used for external validation. An assembly of the CHAOS~\citep{valindria2018multi}, BTCV~\citep{landman2015}, and Pancreas-CT~\citep{TCIA_data} datasets were used for generating synthetic training and validation sets, in which the liver in these datasets is confirmed to be healthy. 
    }\vspace{2px}
    \label{tab:cohort}
    \begin{tabular}{p{0.28\linewidth}P{0.12\linewidth}P{0.1\linewidth}P{0.1\linewidth}P{0.1\linewidth}P{0.12\linewidth}}
    \toprule
    dataset & notation & split & annotation & \# of CTs & tumor \\
    \midrule
\multirow{3}{*}{LiTS~\citep{bilic2019liver}} 
    & cohort 1 & training   & \cmark & 25 & real \\
    & cohort 2 & validation & \cmark & 5 & real  \\
    & cohort 3 & testing    & \cmark & 70 & real \\
    \midrule
\multirow{2}{*}{Assembly~\citep{hu2023label}}
    & cohort 4 & training & \xmark & 25 & synthetic \\
    & cohort 5 & validation & \xmark & 50 & synthetic \\
    \midrule
    
\multirow{2}{*}{FLARE'23~\citep{ma2022fast}} 
    & cohort 6 & validation & \xmark & 50 & synthetic \\
    & cohort 7 & testing & \cmark & 120 & real \\
    \bottomrule
    \end{tabular}
\end{table*}

\tableautorefname~\ref{tab:cohort} summarizes a total of five publicly available datasets used in this study. We group them into three classes.

\begin{itemize}[leftmargin=*]

    \item \textbf{Real-tumor dataset.} We select the LiTS dataset~\citep{bilic2019liver} for training and testing AI models. LiTS provides detailed per-voxel annotations of liver tumors. The tumor types include HCC and secondary liver tumors and metastasis derived from colorectal, breast, and lung cancer. The size of liver tumors ranges from 38mm$^3$ to 349 cm$^3$, and the radius of tumors is approximately in the range of [2, 44] mm. LiTS is partitioned into a training set (\textit{cohort 1}; 25 CT volumes), validation set (\textit{cohort 2}; 5 CT volumes), and test set (\textit{cohort 3}; 70 CT volumes).

    \item \textbf{Healthy CT assembly.}  We have collected a dataset of 75 CT volumes with healthy liver assembled from CHAOS~\citep{valindria2018multi}, Pancreas-CT~\citep{TCIA_data} and BTCV~\citep{landman2015}. This assembled dataset is partitioned into a training set (\textit{cohort 4}; 25 CT volumes) and a validation set (\textit{cohort 5}; 50 CT volumes). As illustrated in \figureautorefname~\ref{fig:methods}(b) For the training set, tumors were dynamically generated within these volumes during training, resulting in a sequential collection of image-label pairs comprising synthetic tumors. For the validation set, we generated three different tumor sizes (small, medium, and large) for each healthy CT volume offline, giving a total of 150 CT volumes.

    \item \textbf{External benchmark.} FLARE'23~\citep{ma2022fast} is used for an external benchmark because it provides out-domain CT volumes from the LiTS dataset. This dataset was specifically chosen due to its extensive coverage, encompassing over 4000 3D CT volumes obtained from more than 30 medical centers. The inclusion of such a diverse dataset ensures the generalizability of the benchmark. The FLARE'23 dataset contains partially labeled annotations. To ensure the suitability of the test set, specific criteria are applied to the annotations. These criteria require that the annotations include per-voxel labeling for both the liver and tumors, with the additional constraint that the connected component of the tumor must intersect with the liver. Adhering to these conditions, we chose the external test set (\textit{cohort 7}; 120 CT volumes). Additionally, same as the assembly dataset, we can use the healthy cases within the FLARE'23 to generate synthetic data to serve as \textit{in-domain} validation set (\textit{cohort 6}; 50 CT volumes), which will be used in \S\ref{sec:indomain_synt_validation}.

\end{itemize}

\subsection{Implementation}\label{sec:implementation}

We have implemented our codes utilizing the MONAI\footnote{\href{https://monai.io/}{https://monai.io/}} framework for the U-Net architecture~\citep{ronneberger2015u}, a well-established network commonly employed in medical image segmentation tasks. During the pre-processing stage, input images undergo clipping with a window range of [-21,189]. Following this, they are normalized to achieve a zero mean and unit standard deviation~\citep{tang2022self}. For training purposes, random patches with dimensions of $96\times96\times96$ are cropped from the 3D image volumes. A base learning rate of 0.0002 is utilized in the training process, accompanied by a batch size of two per GPU. To further enhance the training process, we employ both the linear warmup strategy and the cosine annealing learning rate schedule. Our model is trained for 6,000 epochs, with a model checkpoint being saved every 100 epochs, and a total of 60 model checkpoints are saved throughout the entire training process. During the inference phase, a sliding window strategy with an overlapping area ratio of 0.75 is adopted. To ensure robustness and comprehensiveness in obtaining results, the experiment is conducted ten times each to perform statistical analysis. By averaging all runs, we obtain reliable results. The segmentation performance is evaluated using the Dice Similarity Coefficient (DSC) score, while Sensitivity is used to evaluate the performance of detecting very tiny liver tumors (radius $<$ 5mm).
\section{Result}\label{sec:result}

\textbf{\textit{Summary.}} Using synthetic data as validation can select the best model checkpoint and alleviate the overfitting problem. Furthermore, the AI model developed using our continual learning framework outperforms models trained and validated on a static dataset. The performance is particularly high for detecting small/tiny tumors because we can generate a vast number of examples of small/tiny tumors for both training and validation.

\begin{figure}[t]
    \centering
    \includegraphics[width=\textwidth]{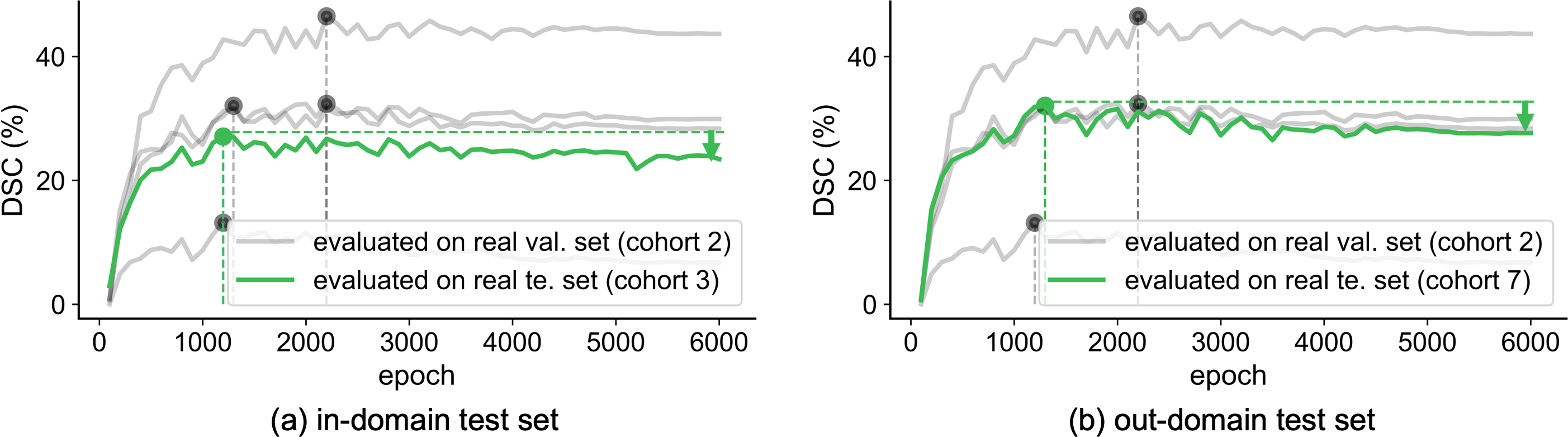}
    \caption{\textbf{The overfitting is due to a small-scale, biased real-tumor validation set.} 
    Model checkpoints are saved at each training epoch, and their performance trajectories are evaluated. The \textcolor{qixin_green}{green} curve plots the test set performance of each checkpoint. It serves as the gold standard for a specific dataset, though the test set performance is often apriori unknown. The \textcolor{mygray}{gray} curve plots the validation set performance of each checkpoint. This performance is accessible during the AI training and instrumental in selecting the best model checkpoint. 
    We train a model on the LiTS training set and subsequently test it on the LiTS test set as in-domain evaluation \textbf{(a)} and the FLARE'23 dataset as out-domain evaluation \textbf{(b)}. At the initial epochs, the model performs increasingly well, but its test set performance declines when trained for more epochs, highlighted by the green arrow. This decline is attributed to \textit{overfitting}, where the model becomes too specialized on the training set and loses its ability to generalize effectively to the test set. The purpose of a validation set is to select the best model checkpoint that is expected to perform well on unseen data. However, in practice, the size and diversity of the validation set may be limited, leading to potential inaccuracies in checkpoint selection. This is evidenced by both in-domain (a) and out-domain (b) evaluations. The dots on the curve represent the best checkpoint selected by the test set (\textcolor{qixin_green}{green}) or the real-tumor validation (\textcolor{mygray}{gray}). A comparative analysis reveals that the checkpoints selected based on the real-tumor validation set might not be the most suitable for test sets.}
    \label{fig:real_validation}
\end{figure}

\subsection{Overfitting is Attributed to Small-Scale, Real-Tumor Validation}\label{sec:overfitting_problem}

To demonstrate the potential limitations of selecting the best model checkpoint based on a small-scale and biased real-tumor validation set, we evaluate all the model checkpoints in real-tumor validation set (cohort 2), in-domain LiTS test set (cohort 3) and out-domain FLARE'23 test set (cohort 7). In-domain test set (cohort 3) assesses the performance of each checkpoint and aids in determining the effectiveness of the selected best checkpoints using the validation set (cohort 2). Out-domain test set (cohort 7) serves as a robust benchmark, providing an enhanced evaluation of the performance of the model checkpoints on out-domain unseen data. 

As shown in \figureautorefname~\ref{fig:real_validation}, two significant observations can be made. Firstly, the best checkpoint identified by the small-scale real-tumor validation set exhibits considerable instability, with notable variations observed when different validation samples are chosen. This result indicates that the small-scale real-tumor validation is inherently biased and lacks the ability to adequately represent the broader range of cases. Secondly, the performance of the best checkpoint determined by the real validation set does not effectively generalize to unseen test data, particularly when confronted with out-domain data. These observations indicate that overfitting can be attributed to a small-scale, biased real-tumor validation set.

\begin{figure}[t]
    \centering
    \includegraphics[width=\textwidth]{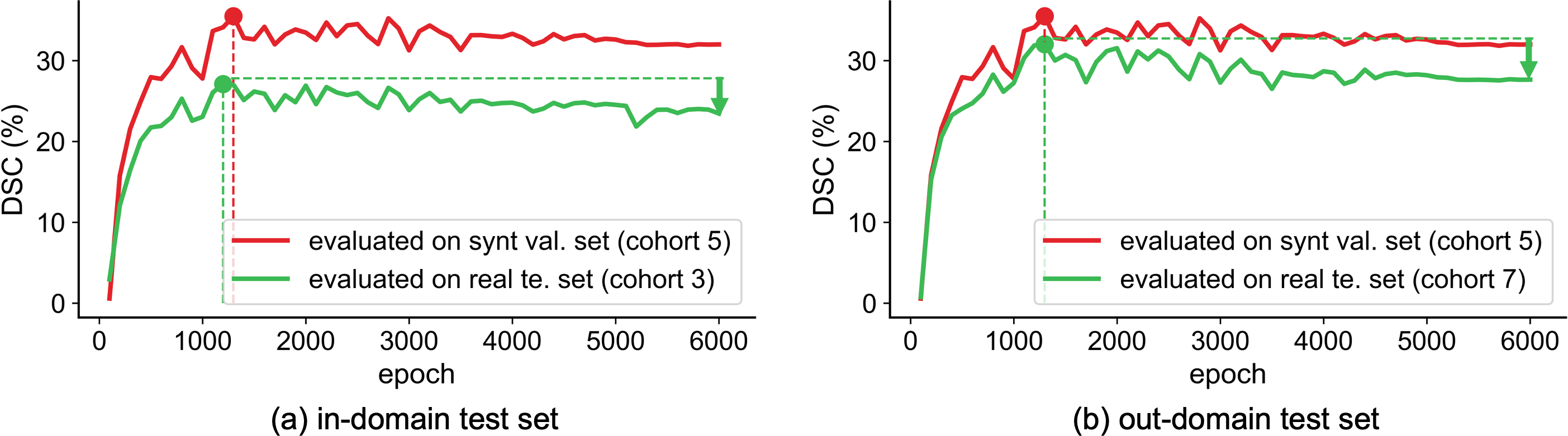}
        \caption{\textbf{The overfitting is alleviated by a large-scale, synthetic-tumor validation set.} Similar to \figureautorefname~\ref{fig:real_validation}, \textbf{(a)} and \textbf{(b)} denote in-domain and out-domain evaluations, and the \textcolor{qixin_green}{green} curves are the test set performance on these two datasets---serving as the gold standard for checkpoint selection. The \textcolor{qixin_red}{red} curves are the validation performance using synthetic tumors (cohort 5). In theory, we can generate an unlimited number of tumors in varied conditions, e.g., size, location, shape, and texture, as needed, using the tumor generator described in \S\ref{sec:tumor_generation}. Such extensive coverage enhances the ability of the validation set to estimate how well the model can be generalized to previously unseen data distributions. As shown in both in-domain and out-domain evaluation, synthetic data as validation can accurately select the best model checkpoint that is almost identical to that selected by test sets.}
    \label{fig:synt_validation}
\end{figure}

\subsection{Overfitting is Alleviated by Large-Scale, Synthetic-Tumor Validation}\label{sec:synthetic_validation}
The overfitting can be alleviated by a diverse, large-scale synthetic-tumor validation set. We conducted a similar experiment to \S~\ref{sec:overfitting_problem}. This experiment involved evaluating all the model checkpoints in synthetic-tumor validation set (cohort 5), in-domain test set (cohort 3) and out-domain test set (cohort 7).

The evaluation trajectory can be observed in \figureautorefname~\ref{fig:synt_validation}, with the synthetic-tumor validation set represented by the red line. It is clear that the best checkpoint selected using the synthetic-tumor validation set performs much better than the best checkpoint chosen using the real-tumor validation set when tested with unseen data. Notably, this improvement is especially remarkable when dealing with out-domain data (cohort 7), as the selected model checkpoint is identical to the one chosen by the out-domain test set. These findings emphasize the effectiveness of synthetic-tumor validation set, which serves as a superior alternative to mitigate overfitting issues.

\begin{figure}[t]
    \centering
    \caption{\textbf{The overfitting can be addressed by continual learning on synthetic data.} We set up the continual learning framework for liver tumor segmentation, described in \S\ref{sec:lifelong_learning}. Its learning curves are presented in \textcolor{qixin_red}{red}, referred to as \textit{dynamic training}. In comparison, \textcolor{mygray}{gray} curves are the conventional framework training with a limited number of real data, referred to as \textit{static training}. Both in-domain \textbf{(a)} and out-domain \textbf{(b)} evaluations show that the AI model continuously trained on synthetic data outperforms the one trained on real data.}
    \includegraphics[width=\textwidth]{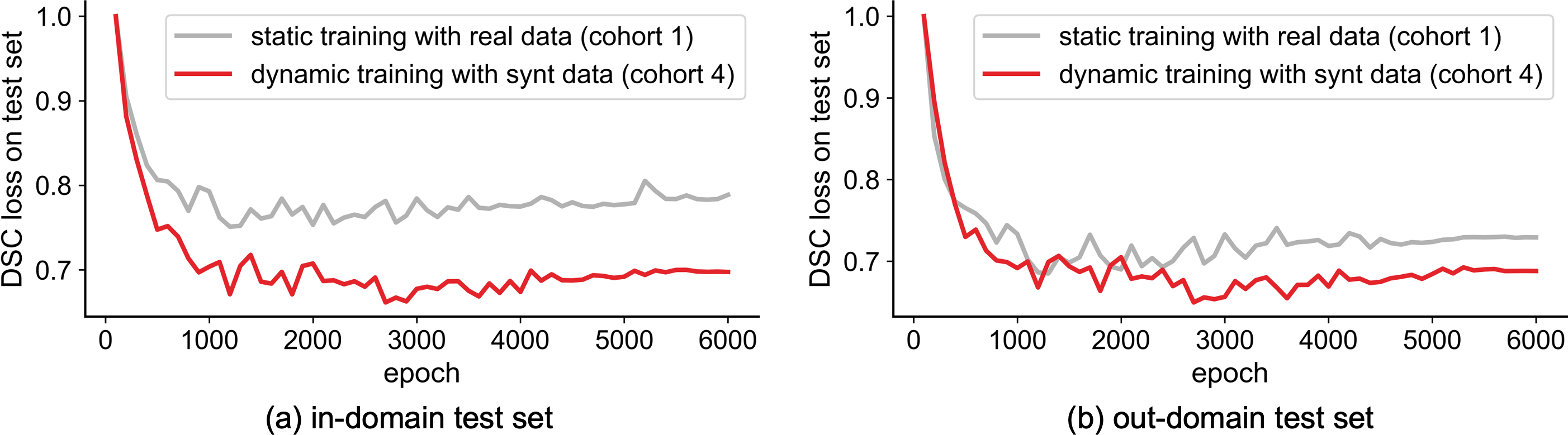} 
\label{fig:real_synt_training}
\end{figure}

\begin{table*}[t]
    \footnotesize
    \centering
    \begin{tabular}{p{0.12\linewidth}|P{0.18\linewidth}P{0.18\linewidth}|P{0.18\linewidth}P{0.18\linewidth}}
    \toprule
 & \multicolumn{2}{c|}{evaluated on real data} & \multicolumn{2}{c}{evaluated on synthetic data} \\
    \cline{2-5}
     & train $@$ real & train $@$ synt & train $@$ real & train $@$ synt \\
    \midrule
    cohort 3 & 26.7 (22.6-30.9)  & \textbf{33.4 (28.7-38.0)} & 27.0 (23.7-30.3)  & \textbf{34.5 (30.8-38.2)} \\
    \midrule
    cohort 7 & 31.1 (26.0-36.2)  & \textbf{33.3 (30.6-36.0)} & 32.0 (28.5-35.5)  & \textbf{35.4 (32.1-38.7)} \\ 
    \bottomrule
    \end{tabular}

    \caption{
        \textbf{
        Synthetic data for both training and validation.} As shown in \figureautorefname~\ref{fig:real_synt_training}, AI model can be trained on either static real data or dynamic synthetic data. The terms ``train $@$ real'' and ``trained $@$ synt'' denote static training with real data and dynamic training with synthetic data, respectively. We save the model checkpoints at each training epoch and then use a validation set to select the \textit{best} model. These selected model checkpoints are tested on the in-domain LiTS test set (cohort 3) and out-domain FLARE'23 test set (cohort 7). We report the DSC score (\%) and 95\% confidence interval achieved on the test set. The result reveals that training and validating AI models with our continual learning framework can significantly improve liver tumor segmentation.
    }\vspace{2px}
\label{tab:main_result}
\end{table*}

\subsection{Overfitting Can Be Addressed by Continual Learning on Synthetic Data}\label{sec:synthetic_training}

We have shown the effectiveness of large-scale synthetic-tumor validation set. 
Now, we will shift our attention to synthetic data in handling the overfitting problem from a training perspective. For this purpose, we introduce a continual learning framework on synthetic data, detailed in \S~\ref{sec:lifelong_learning}. 

The training trajectory of static training on real data and dynamic training on synthetic data is shown in \figureautorefname~\ref{fig:real_synt_training}, and the liver tumor segmentation results are presented in \tableautorefname~\ref{tab:main_result}. Specifically, the AI model trained on static real data demonstrates a DSC score of 26.7\% for the in-domain test set (cohort 3) and 31.1\% for the out-domain test set (cohort 7). In comparison, the AI model developed using our continual learning framework with synthetic data achieves notably higher DSC scores, reaching 34.5\% on cohort 3 and 35.4\% on cohort 7, respectively. These results indicate a notable improvement in the synthetic data. Based on these findings, we can confidently assert that incorporating our continual learning framework with synthetic data allows us to effectively address the issue of overfitting, encompassing both the training and validation perspectives.

\begin{figure}[t]
    \centering
    \includegraphics[width=0.8\textwidth]{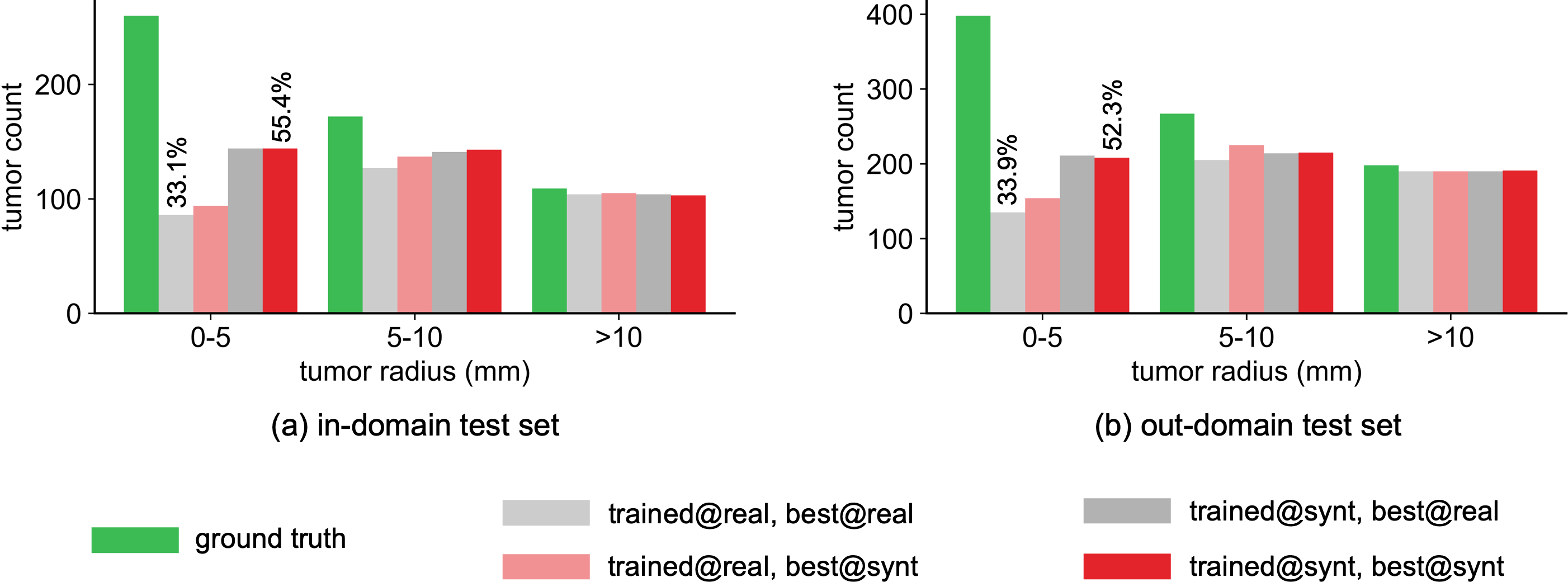} 
    \caption{\textbf{Synthetic data can benefit early cancer detection.} We evaluate the efficacy of synthetic data in detecting tiny liver tumors (radius $<$ 5mm). Specifically, the AI model developed with our continual learning framework -- trained and evaluated with synthetic data -- is evaluated on tumor detection. We compare it with AI model developed on static real data. The ``trained$@$real`` and ``trained$@$synt`` hold the same meaning as described in \tableautorefname~\ref{tab:main_result}. The ``best$@$real`` and ``best$@$synt`` denote the selection of best checkpoints based on real- and synthetic- tumor validation sets, respectively. We report the sensitivities (\%) achieved on the test set. As shown in both in-domain \textbf{(a)} and out-domain \textbf{(b)} evaluations, our continual learning framework with synthetic data proves to be effective in detecting tiny liver tumors (radius $<$ 5mm), thereby benefiting early cancer detection.} 
\label{fig:small_tumor_detection}
\end{figure}
\subsection{Synthetic Data Can Benefit Early Cancer Detection}\label{sec:small_tumor_detection}

Early tumor (radius $<$ 5mm) detection plays a critical role in clinical applications, providing valuable information for early cancer diagnosis. Acquiring real data of such a small size is challenging, often posing difficulties or even making it impossible to acquire them. However, our strategy can dynamically generate numerous tiny tumors as required. As a result, the AI model developed within the continual learning framework yields a significant improvement in detecting tiny liver tumors.
The improvement can be found in \figureautorefname~\ref{fig:small_tumor_detection}. We assessed the sensitivity of the AI model under different settings. The performance of the AI model trained and validated on the static real data is 33.1\% for the in-domain test set (cohort 3) and 33.9\% for the out-domain test set (cohort 7). Comparatively, the AI model developed using our continual learning framework on synthetic data gives a sensitivity of 55.4\% for cohort 3 and 52.3\% for cohort 7. These results prove the effectiveness of our framework in early detection of cancer.

\begin{figure}[t]
    \centering
    \includegraphics[width=0.6\textwidth]{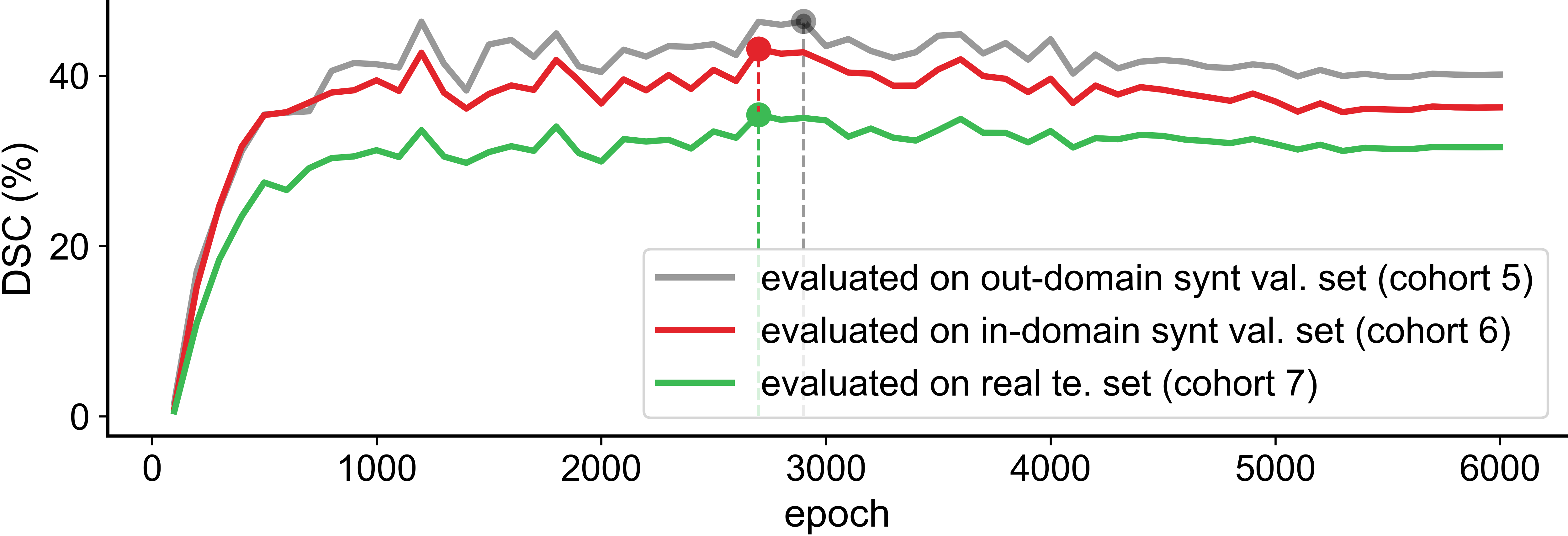}
    \caption{\textbf{The continual learning framework is enhanced by in-domain synthetic-tumor validation.} Our framework enables the utilization of healthy cases from various domains. Specifically, an interesting situation arises where we can directly create synthetic-tumor validation using healthy cases from the same domain as the test set. This in-domain synthetic-tumor validation set can benefit our continual learning framework. To illustrate this advantage, we utilize the FLARE'23 dataset, which contains both disease cases and healthy cases. We trained the model using dynamic synthetic data and saved model checkpoints at each training epoch. Subsequently, we evaluated them on three datasets, i.e., out-domain synthetic-tumor validation set from assembly dataset (cohort 5, \textcolor{gray}{gray} curve), in-domain synthetic-tumor validation set from FLARE'23 dataset (cohort 6, \textcolor{qixin_red}{red} curve), and FLARE'23 test set served as the gold standard (cohort 7, \textcolor{qixin_green}{green} curve). As shown, the in-domain synthetic-tumor validation set accurately identifies the best model, which aligns with the model selected by the FLARE'23 test set.}
    \label{fig:indomain_synt_validation}
\end{figure}

\subsection{Continual Learning Framework is Enhanced by In-domain Synthetic-Tumor Validation}\label{sec:indomain_synt_validation}

We have demonstrated the effectiveness of our continual learning framework. Moving forward, let's consider the framework itself. Synthetic data offers a significant advantage as we can utilize healthy CT volumes from various domains. A notable scenario arises wherein we can directly generate synthetic-tumor validation using healthy cases from the same domain as the test set, providing valuable insights for our continual learning framework. As shown in \figureautorefname~\ref{fig:indomain_synt_validation}, the in-domain synthetic-tumor validation set showcases its capability to accurately identify the best model, which aligns with the model selected by the test set. This result highlights that the continual learning framework yields more favorable outcomes when we can generate in-domain synthetic-tumor validation.

\section{Conclusion}\label{sec:conclusion}

Data synthesis strategies continue to pique the interest of researchers and practitioners, propelling ongoing investigations within this field. This paper justifies the potential and stresses the necessity of leveraging synthetic data as validation to select the best model checkpoint along the training trajectory. Moreover, by employing a continual learning framework on synthetic data, we realize a marked improvement in liver tumor segmentation as well as in the early detection of cancerous tumors compared with the static training on real data, where procuring ample annotated examples can be cost-prohibitive. It is particularly valuable in scenarios characterized by limited annotated data. In the future, we plan to improve the generation of synthetic tumors and verify our findings across different organs, such as the pancreas, kidneys, and stomach.

\subsubsection*{Acknowledgments}
This work was supported by the Lustgarten Foundation for Pancreatic Cancer Research and the Patrick J. McGovern Foundation Award. We appreciate the effort of the MONAI Team to provide open-source code for the community. This work has partially utilized the GPUs provided by ASU Research Computing and NVIDIA. We thank Yu-Cheng Chou, Junfei Xiao, Xiaoxi Chen, and Bowen Li for their constructive suggestions at several stages of the project.

\clearpage
\bibliographystyle{acl_natbib}
\bibliography{refs,zzhou}

\appendix

\clearpage
\section{Distribution of Real Tumors}\label{appdix:real_tumor_distribution}
\begin{figure}[h]
    \centering
    \includegraphics[width=0.55\textwidth]{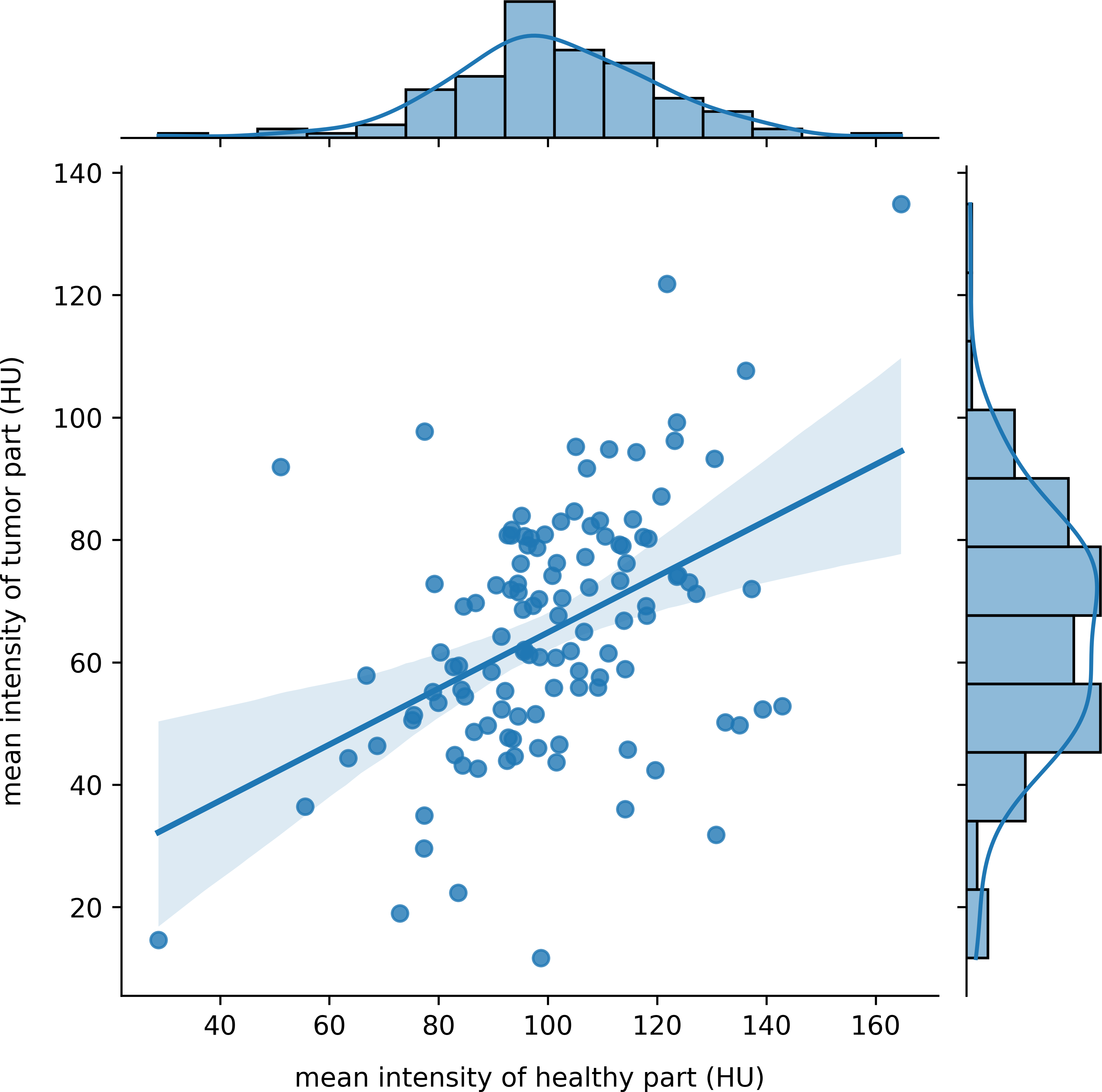}
        \caption{\textbf{Intensity distribution of liver tumors and their healthy counterparts.} Our tumor generator, which is based on modeling and medical knowledge, incorporates the distributional characteristics of real tumors. We present the intensity distributions of real tumors obtained from the LiTS dataset. }
    \label{fig:appendix_distribution}
\end{figure}
\begin{figure}[h]
    \centering
    \includegraphics[width=0.55\textwidth]{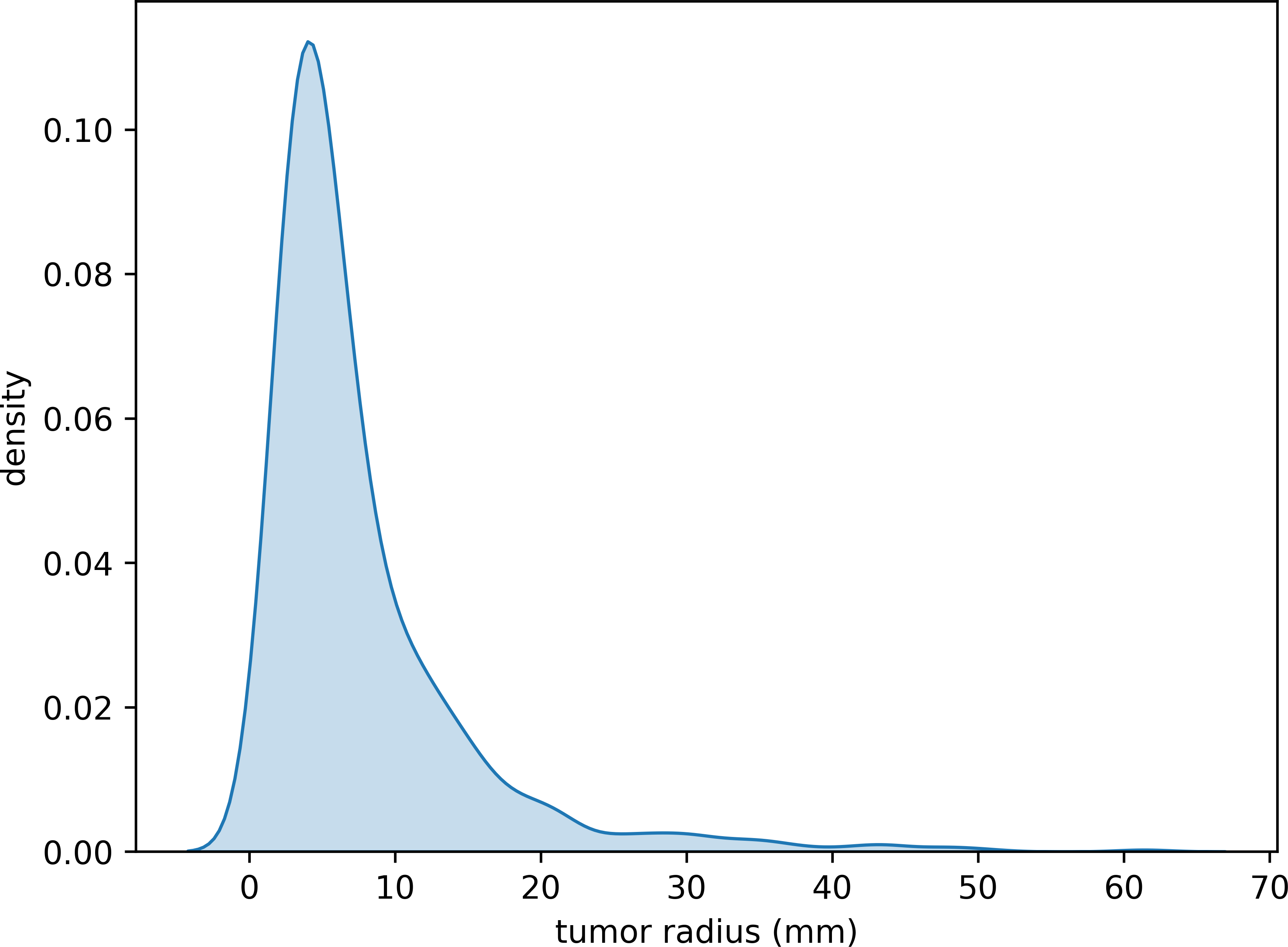}
        \caption{\textbf{Size distribution of liver tumors.} We have calculated the size distribution of liver tumors from the LiTS dataset. This tumor size distribution will serve as a guide for determining the sizes of the synthetic tumors we generate.}
    \label{fig:appendix_size_dis}
\end{figure}

\clearpage
\section{Shape Examples}\label{appdix:shape_examples}
\begin{figure}[h]
    \centering
    \includegraphics[width=0.65\textwidth]{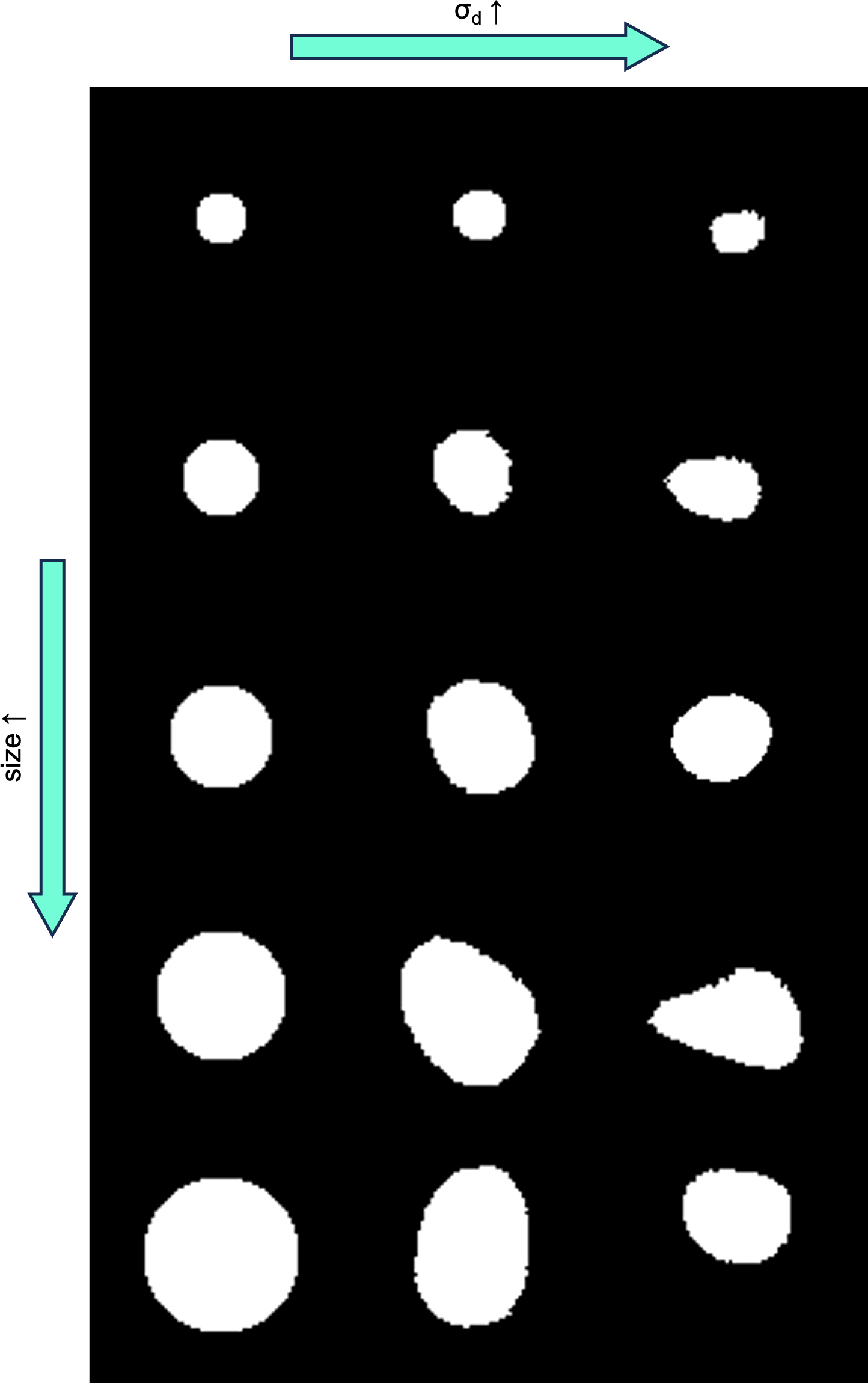}
        \caption{\textbf{Examples of the generated shape.} The tumor generator pipeline enables us to control the size and deformation of the generated tumors. Here, we present some examples of generated shapes under different conditions.}
    \label{fig:appendix_shape_examples}
\end{figure}

\newpage
\section{Texture Examples}\label{appdix:texture_examples}
\begin{figure}[h]
    \centering
    \includegraphics[width=0.65\textwidth]{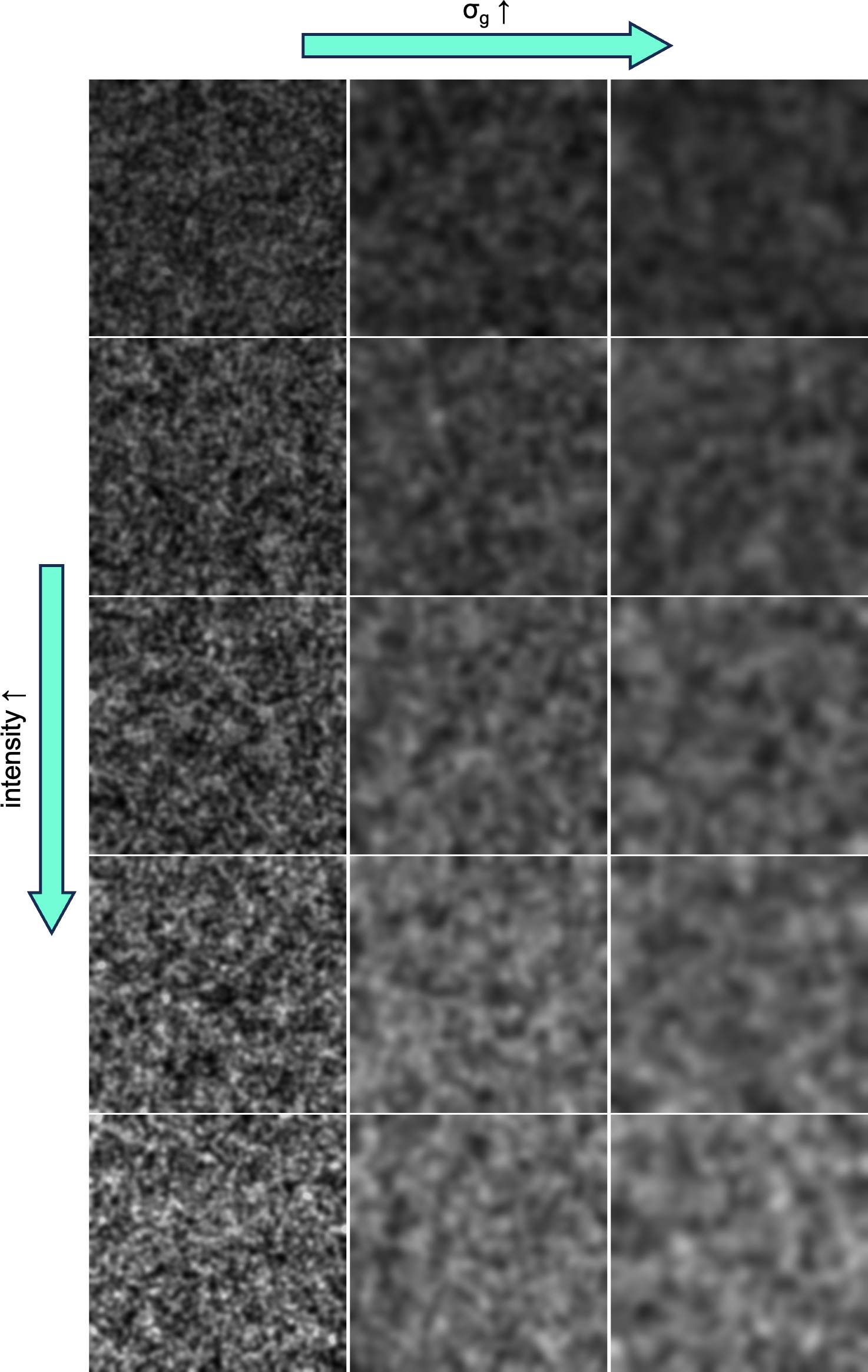}
        \caption{\textbf{Examples of the generated texture.} Our data synthesis strategy also enables us to generate different textures, as illustrated here for visualization.}
    \label{fig:appendix_texture_examples}
\end{figure}

\clearpage
\section{Synthetic tumor examples}\label{appdix:synthetic_examples}
\begin{figure}[h]
    \centering
    \includegraphics[width=\textwidth]{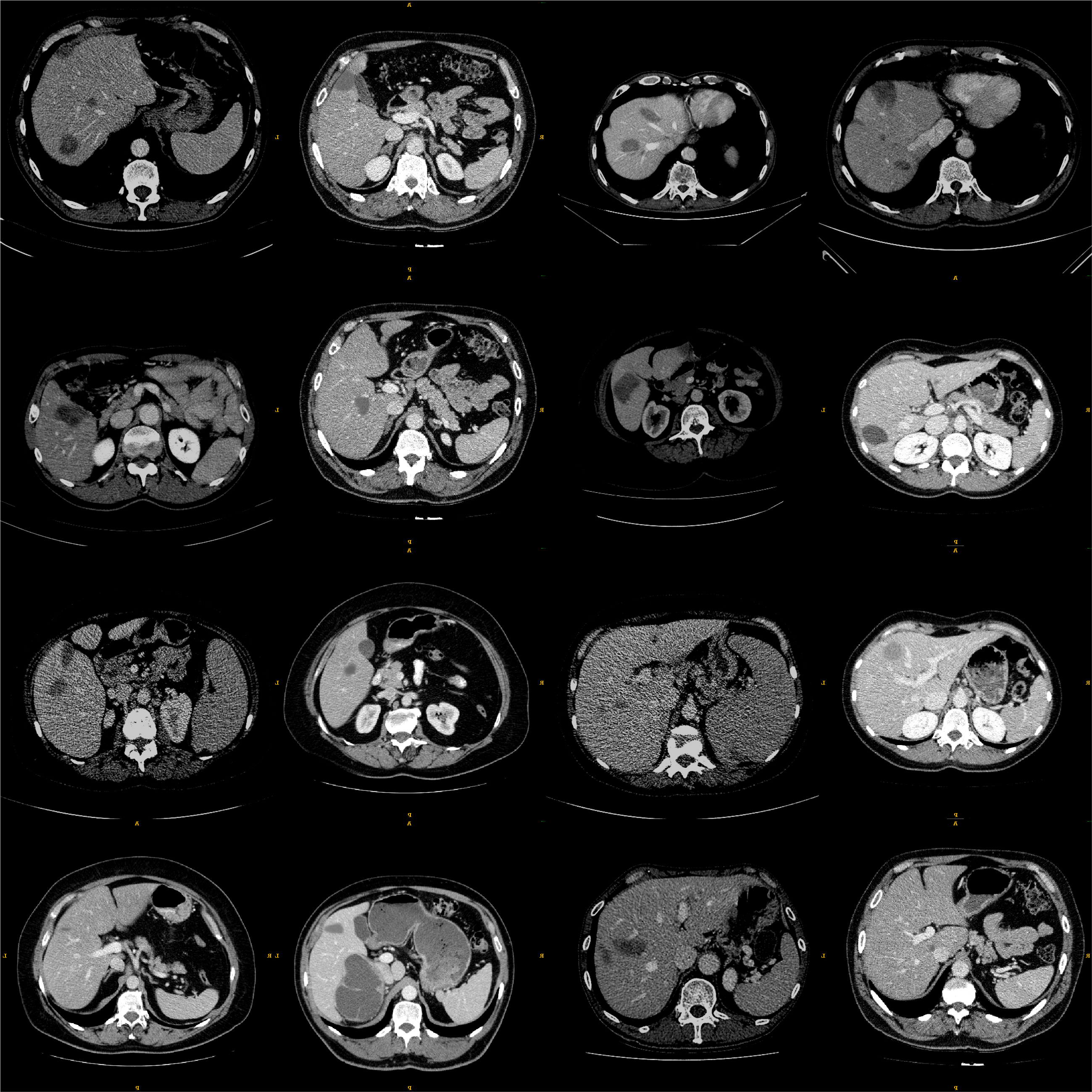}
        \caption{\textbf{Visualization of synthetic data.} By combining all the pipelines together, we can obtain a wide range of diverse synthetic data for validation and training.}
    \label{fig:appdix_examples}
\end{figure}

\end{document}